\documentclass[11pt,a4paper]{article}
\usepackage[hyperref]{acl2020}
\usepackage{times}
\usepackage{latexsym}

\usepackage{colortbl}
\usepackage{booktabs}
\usepackage{multirow}
\usepackage{graphicx}
\usepackage[T1]{fontenc}

\usepackage{amsmath, amsthm, amssymb}

\usepackage{microtype}

\aclfinalcopy 
\def\aclpaperid{2336} 

\newcommand{\inc}[2]{\cellcolor{cyan!#1}#2}
\newcommand{\dec}[2]{\cellcolor{red!#1}#2}
\newcommand{\token}[1]{\texttt{#1}}

\definecolor{hred}{RGB}{247,202,197}
\definecolor{horange}{RGB}{242,217,198}
\definecolor{hyellow}{RGB}{247,225,147}
\definecolor{hlgreen}{RGB}{200,207,141}
\definecolor{hdgreen}{RGB}{172,215,156}
\definecolor{hlblue}{RGB}{181,228,227}
\definecolor{hdblue}{RGB}{188,220,246}
\definecolor{hpurple}{RGB}{219,218,246}

\newcommand{\aggressiveness}{\colorbox{hred}{\textsc{aggressiveness}}}
\newcommand{\optimism}{\colorbox{horange}{\textsc{optimism}}}
\newcommand{\love}{\colorbox{hyellow}{\textsc{love}}}
\newcommand{\submission}{\colorbox{hlgreen}{\textsc{submission}}}
\newcommand{\awe}{\colorbox{hdgreen}{\textsc{awe}}}
\newcommand{\disapproval}{\colorbox{hlblue}{\textsc{disapproval}}}
\newcommand{\remorse}{\colorbox{hdblue}{\textsc{remorse}}}
\newcommand{\contempt}{\colorbox{hpurple}{\textsc{contempt}}}
\newcommand{\trust}{\colorbox{hlgreen}{\textsc{trust}}}
\newcommand{\admiration}{\colorbox{hlgreen}{\textsc{admiration}}}

\newcommand{\fear}{\colorbox{hdgreen}{\textsc{fear}}}
\newcommand{\loathing}{\colorbox{hpurple}{\textsc{loathing}}}
\newcommand{\joy}{\colorbox{hyellow}{\textsc{joy}}}
\newcommand{\ecstasy}{\colorbox{hyellow}{\textsc{ecstasy}}}
\newcommand{\grief}{\colorbox{hlblue}{\textsc{grief}}}

\newcommand{\anger}{\colorbox{hred}{\textsc{anger}}}
\newcommand{\annoyance}{\colorbox{hred}{\textsc{annoyance}}}

\newcommand{\sepr}{\hspace{0.2em}}

\title{Detecting Perceived Emotions in Hurricane Disasters}

 \author{
    \textbf{Shrey Desai}$^1$\quad\textbf{Cornelia Caragea}$^2$\quad\textbf{Junyi Jessy Li}$^1$\\
    $^1$The University of Texas at Austin\quad $^2$University of Illinois at Chicago\\
    \texttt{\{shreydesai@, jessy@austin.\}utexas.edu}\quad\texttt{cornelia@uic.edu}
}

\date{}

\begin{document}

\maketitle
\begin{abstract}
Natural disasters (e.g., hurricanes) affect millions of people each year, causing widespread destruction in their wake. People have recently taken to social media websites (e.g., Twitter) to share their sentiments and feelings with the larger community. Consequently, these platforms have become instrumental in understanding and perceiving emotions at scale. In this paper, we introduce \textsc{HurricaneEmo}, an emotion dataset of 15,000 English tweets spanning three hurricanes: Harvey, Irma, and Maria. We present a comprehensive study of fine-grained emotions and propose classification tasks to discriminate between coarse-grained emotion groups. Our best BERT \citep{devlin-etal-2019-bert} model, even after task-guided pre-training which leverages unlabeled Twitter data, achieves only 68\% accuracy (averaged across all groups). \textsc{HurricaneEmo} serves not only as a challenging benchmark for models but also as a valuable resource for analyzing emotions in disaster-centric domains.
\end{abstract}

\section{Introduction}

Natural disasters cause thousands of deaths and displace hundreds of millions each year \citep{ritchie2014natural}. These catastrophic events not only induce material destruction but also stir an integral part of being human: our emotions. Disasters adversely affect individuals' mental states \citep{fritzdisaster,kinston1974disaster}, and therefore it is no surprise that many take to social media (e.g., Twitter) to share their feelings. Social media websites, as a result, have become an essential platform for understanding the expression and perception of emotions at a significantly larger scale \cite{mohammad-2012-emotional,Wang:2012:HTB:2411131.2411712,mohammad2015using,volkova-bachrach-2016-inferring,abdul-mageed-ungar-2017-emonet}, with far reaching potential influences from academic research to public policy \citep{dennis2006making,fritze2008hope,fraustino2012social}.

While natural language processing methods have been effective for emotion detection \citep{strapparava-mihalcea-2007-semeval}, existing resources struggle in disaster-centric domains, in part due to distributional shifts. Emotion detection in natural disasters (e.g., hurricanes) requires implicit reasoning not available as surface-level lexical information. For example, in \emph{``of course, [we]$_1$ still have the [storm surge]$_2$ coming,''} given the context, we can reasonably infer discontent towards the ``storm surge'' despite the absence of polarizing words. Therefore, distantly supervised techniques largely based on lexical units \citep{mohammad2013saif,abdul-mageed-ungar-2017-emonet} fail to capture this type of deeper semantic phenomena.

Our paper presents a comprehensive investigation into perceived emotions in hurricane disasters. To this end, we introduce \textsc{HurricaneEmo}, a dataset of 15,000 disaster-related tweets (in English) streamed during Hurricanes Harvey, Irma, and Maria, which were devastating tropical storms occurring in the 2017 Atlantic hurricane season \citep{belles2017hurricane}. Our samples are annotated with fine-grained emotions derived from the Plutchik Wheel of Emotions \cite{plutchik2001the}, a well-defined ontology of emotion classes commonly used in computational social science \cite{abdul-mageed-ungar-2017-emonet}.\footnote{Specifically, we use Plutchik-8 and Plutchik-24 emotions. We refer readers to \citet{plutchik2001the} for an in-depth discussion on their conception.} To measure inter-annotator agreement on fine-grained emotion labels, we conceptualize the \textbf{P}lutchik \textbf{E}motion \textbf{A}greement (PEA) metric (\S\ref{sec:dataset-construction}). PEA is intuitively grounded; our human evaluation shows workers agree with PEA's rankings 88\% of the time. Furthermore, we perform insightful analyses on \textit{implicit} and \textit{explicit} emotions in hurricane tweets (\S\ref{sec:qualitative-analysis}). Quite surprisingly, we find consistencies in Plutchik-24 emotion distributions across Hurricanes Harvey, Irma, and Maria.

\textsc{HurricaneEmo} also serves as a challenging new benchmark for large-scale, pre-trained language models. We establish baselines for a coarser Plutchik-8 emotion detection task using BERT \citep{devlin-etal-2019-bert} and RoBERTa \citep{liu2019roberta} (\S\ref{sec:baseline-modeling}). Our experiments reveal: (1) BERT only achieves 64\% (averaged) accuracy; and (2) using ``better'' pre-trained models (e.g., RoBERTa) does not help, which is a strikingly different trend than most leaderboards \citep{wang-etal-2018-glue}. To better understand their pitfalls, in particular BERT, we conduct a comprehensive error analysis of 200 incorrectly predicted samples. In addition, we incorporate stronger inductive biases into BERT via pre-training on related tasks, which culminates in (averaged, absolute) +4\% accuracy (\S\ref{sec:task-guided-pretraining}). Finally, we propose unsupervised domain adaptation to bridge the domain gap between existing large-scale emotion datasets (e.g., \textsc{EmoNet} \citep{abdul-mageed-ungar-2017-emonet}) and \textsc{HurricaneEmo} (\S\ref{sec:fine-grained-uda}). Our code and datasets are made publicly available.\footnote{\url{https://github.com/shreydesai/hurricane}}

\section{Related Work}
\label{sec:related-work}

Emotion detection has been extensively studied in news headlines \cite{strapparava-mihalcea-2007-semeval,katz2007swat}, blog posts \cite{aman07}, health-related posts \cite{khanpour-caragea-2018-fine}, and song lyrics \cite{strapparava-etal-2012-parallel}, but only recently, in social media websites (e.g., Twitter, Facebook) \cite{mohammad-2012-emotional,Wang:2012:HTB:2411131.2411712,mohammad2015using,volkova-bachrach-2016-inferring,abdul-mageed-ungar-2017-emonet}. However, emotion detection in disaster-centric domains, despite its practical importance, is limited. \citet{Schulz2013AFS} (singlehandedly) annotate 2,200 Hurricane Sandy tweets using Ekman-6 emotions \cite{ekman1992an}. In contrast, we introduce 15,000 annotated tweets from multiple hurricanes with (much more fine-grained) Plutchik-24 emotions. Unlike \citet{abdul-mageed-ungar-2017-emonet}, we focus on readers' \textit{perceived} emotions rather than writers' \textit{intended} emotions. 

Furthermore, in disaster-centric domains, the lack of labeled data required to train reliable models precludes the use of supervised learning techniques. Several works propose to use labeled data from prior (source) disasters to learn classifiers for new  (target) disasters \citep{Verma2011ICWSM,Nguyen2017ICWSM,Imran:2013:PED:2487788.2488109,ImranMS16iscram,Caragea2016IdentifyingIM}. However, due to the unique nature of each disaster (e.g., type, geographical location, season, cultural differences among the affected population), the source disaster may not accurately reflect the characteristics of the target disaster \citep{Palen2016science,Imran:2015:PSM:2775083.2771588}. Domain adaptation techniques address these challenges by efficiently using large amounts of unlabeled target domain data, consequently outperforming the aforementioned supervised techniques \citep{alam-etal-2018-domain,Li2017TowardsPU}. Our work contributes to disaster-centric emotion detection in three ways by: (1) introducing a dataset large enough to train supervised classifiers; (2) exploring various forms of pre-training to instill strong inductive biases; and (3) establishing domain adaptation baselines by leveraging emotive samples obtainable via distant supervision.

\section{Dataset Construction}
\label{sec:dataset-construction}

In this section, we present \textsc{HurricaneEmo}, an annotated dataset of 15,000 English tweets from Hurricanes Harvey, Irma, and Maria. We detail each component, including the initial preprocessing (\S\ref{sec:preprocessing}), annotation procedures (\S\ref{sec:annotation}), and the formulation and calculation of inter-annotator agreement (\S\ref{sec:inter-annotator-agreement}).

\subsection{Preprocessing}
\label{sec:preprocessing}

\citet{raychowdhury2019keyphrase} release a repository of large-scale Twitter datasets consisting of tweets streamed during the Harvey, Irma, and Maria hurricanes, which we will refer to as \textsc{HurricaneExt} (i.e., extended). We use their tweets as a starting point for the construction of our dataset. We perform two types of preprocessing. First, we replace usernames and links with \token{<USER>} and \token{<URL>}, respectively, then eliminate duplicate tweets. Second, we use filtering techniques to ensure the resulting tweets contain emotive content.

We assume a lexical prior over emotion tweets, that is, requiring that an emotive tweet consist of \textit{at least} one word derived from \textsc{EmoLex} \cite{mohammad2013saif}.  \textsc{EmoLex} consists of 14,182 crowdsourced words associated with several emotion categories. Critically, these words appear in emotional contexts, but are not necessarily emotion words themselves. For example, ``payback'' is related to the emotion ``anger,'' but is also used extensively in finance. Significant past work \cite{BravoMarquez2014MetalevelSM,majumder2017,Giatsoglou2017SentimentAL} has used this lexicon to bootstrap their emotion datasets, since the alternatives are (1) using unlabeled tweets as-is or (2) using a model to classify emotional tweets. Initially, we started with (1) and did no emotion-related preprocessing. However, the dataset contained many spurious tweets, such as snippets of news articles, that had little to do with emotions. The level of noise rendered the data prohibitively costly to annotate. For (2), there is simply no such large-scale data to train on, and existing resources like \textsc{EmoNet} manifest an even stronger prior where tweets are only included if they explicitly contain an emotion hashtag (e.g., \textit{\#sad}, \textit{\#angry}, \textit{\#happy}).

\subsection{Annotation}
\label{sec:annotation}

We randomly sample 5,000 tweets each for annotation from the filtered datasets for Harvey, Irma, and Maria; in total, this yields 15,000 annotations. We request workers on Amazon Mechanical Turk to label tweets with a list of Plutchik-24 emotions. Furthermore, to enable fine-grained emotion analysis, we do not crowdsource Plutchik-8 emotions directly. We require that workers reside in the US and have completed 500+ HITs with an acceptance rate $\ge$ 95\%. Each HIT is completed by 5 workers.

\subsection{Inter-Annotator Agreement}
\label{sec:inter-annotator-agreement}

In this section, we elaborate on our PEA metric for computing inter-annotator agreement with fine-grained emotion labels.

\paragraph{Challenges.} Fine-grained emotion annotation presents several challenges for evaluating inter-annotator agreement. First, because a tweet can convey multiple emotions, we allow workers to select more than one Plutchik-24 emotion. This implies an agreement  metric must support scoring \textit{sets} of categorical values. \citet{passonneau-2004-computing} use set distance metrics for capturing agreement between coreference cluster annotations. Similarly, \citet{wood-etal-2018-comparison} incorporate Jaccard's similarity in Krippendorff's alpha. However, these methods would penalize fine-grained emotions equally, which is not ideal. For the Plutchik wheel, the proximity of any two emotions represents their relatedness. For example, \trust\sepr and \admiration\sepr belong to the same emotion group while \loathing\sepr and \admiration\sepr are orthogonal to each other.

\paragraph{PEA Scores.} We introduce the \textbf{P}lutchik \textbf{E}motion \textbf{A}greement---hereafter referred to as PEA---to address these challenges. We superimpose a unit circle onto the Plutchik wheel, representing each Plutchik-8 emotion as a polar coordinate (e.g., \disapproval\sepr $ = (\frac{\sqrt{2}}{2}, \frac{-\sqrt{2}}{2})$). Intuitively, the angles between Plutchik-8 emotions represent how similar or dissimilar they are. If two Plutchik-24 annotations belong to the same Plutchik-8 group, we do not penalize them (e.g., \joy\sepr and \ecstasy\sepr incur no penalty). Otherwise, we enforce a linear penalty based on how radially separate the annotations are (e.g., \ecstasy\sepr and \grief\sepr incur the highest penalty). Higher PEA scores imply more agreement.

\begin{figure}[t!]
    \centering
    \includegraphics[width=4cm]{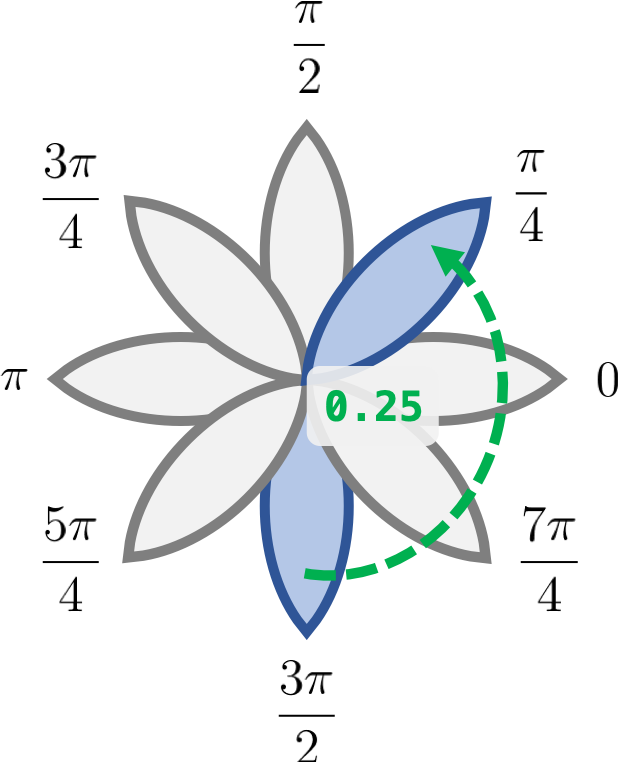}
    \caption{Visualization of the PEA metric. The unit circle is superimposed on the Plutchik wheel, and each Plutchik-8 emotion is assigned a radian value. In this example, the (normalized) distance between the emotions corresponding to $\frac{3\pi}{2}$ and $\frac{\pi}{4}$ is 0.25.}
    \label{fig:appendix-pea-score}
\end{figure}

\paragraph{Example.} Figure \ref{fig:appendix-pea-score} visualizes our metric. In this example, two annotators select emotions with radians $\frac{3\pi}{2}$ and $\frac{\pi}{4}$, respectively. The $|f(e_x^{(i)}) - f(e_y^{(j)})|$ term evaluates to $\frac{5\pi}{4}$. Then, it is normalized using $\frac{1}{\pi}$, yielding $\frac{5}{4} = 1.25$. Finally, we subtract to obtain the agreement score: $|1-1.25|=0.25$. Intuitively, this makes sense as the decisions are only slightly better than being in complete disagreement (i.e., orthogonal).

\paragraph{Formulation.} For clarity, we introduce notation. Let $w_x$ and $w_y$ denote workers with (categorical) annotation sets $\{e^{(i)}_x\}^n_{i=1}$ and $\{e^{(j)}_y\}^m_{j=1}$, respectively. The pairwise agreement $d(w_x, w_y)$ between the workers is computed as:
\begin{align*}
    \frac{1}{n}\sum_{i=1}^n \max_j \big( |1 - \frac{1}{\pi}|f(e_x^{(i)}) - f(e_y^{(j)})|| \big)
\end{align*}
where $\frac{1}{\pi}$ is a normalizing constant and $f: \Omega \rightarrow \mathbb{R}$ is a map from Plutchik-8 emotions to radians. Given a collection of workers that annotated a tweet, we obtain per-worker PEA scores by averaging over all possible pairwise agreements. For example, if workers $w_{1-3}$ annotated the same tweet, PEA($w_1$) $= \frac{1}{2}(d(w_1, w_2) + d(w_1, w_3))$. For quality control, we filter annotations from workers with PEA $\le$ 0.55. This threshold is determined through manual inspection of 50 workers and their annotations. The (averaged, per-worker) PEA scores for each hurricane are: Harvey (65.7), Maria (67.3), and Irma (70.3).\footnote{A reasonable interpretation of PEA scores may be as follows: 0---25 (no agreement), 25---50 (poor agreement), 50---75 (moderate agreement), 75---100 (high agreement).}

\paragraph{Human Evaluation.} We perform a human evaluation with our proposed metric, which is absent in previous work for measuring inter-annotator agreement for emotion annotations \cite{wood-etal-2018-comparison,ohman-etal-2018-creating}. Crowdsourced workers are asked to determine the agreement between two annotation pairs constructed from three annotators, that is, A: $(e_1, e_2)$ and B: $(e_1, e_3)$. They choose between three options: (1) A has higher agreement than B; (2) A and B have (roughly) the same agreement; and (3) B has higher agreement than A. 88.2\% of the worker rankings match with PEA's rankings, pointing towards strong human agreement. The workers themselves in this study also show good agreement according to Krippendorff's alpha ($\alpha$ = 74.0) \cite{artstein2008inter}.\footnote{See Appendix \ref{sec:appendix-plutchik-emotion-agreement} for details on our procedures.}

\section{Qualitative Analysis}
\label{sec:qualitative-analysis}

\subsection{Dataset Overview}
\label{sec:dataset-overview}

\begin{table}[t!]
\small
\centering
\begin{tabular}{lrrrrr}
\toprule
 \multicolumn{1}{c}{} & \multicolumn{2}{c}{Vocabulary} & \multicolumn{3}{c}{Features (\%)} \\
    \cmidrule(r){2-3} \cmidrule(r){4-6}
 Hurricane & Orig. & Filt. & \# & @ & // \\
\midrule
Harvey & 20.6 K & 14.4 K & 48.1 & 27.4 & 85.3 \\
Irma & 14.6 K & 8.8 K & 41.4 & 22.5 & 81.7 \\
Maria & 21.6 K & 15.8 K & 36.5 & 30.3 & 78.3 \\
\bottomrule
\end{tabular}
\caption{Per-hurricane dataset statistics. In the vocabulary section, Orig. shows vocabulary counts (obtained through whitespace tokenization) and Filt. shows counts after \token{<USER>} and \token{<URL>} preprocessing. In the features section, we show the percentage of tweets with hashtags (\#), user mentions (@), and links (//).}
\label{tab:dataset_stats}
\end{table}

\begin{table}
\small
\centering
\begin{tabular}{p{4.75cm}p{2cm}}
\toprule
Mexico helped us during Houston, lets return the favor! & joy, admiration, pensiveness \\
\midrule
Hurricane Irma is hitting Florida. Everyone evacuated Here I am, still in Florida bring it on Irma, bring it on. & acceptance, anticipation, vigilance \\
\midrule
puerto rico should be the ONLY THING in American News. \token{<URL>} & anger, annoyance, interest \\
\bottomrule
\end{tabular}
\caption{Samples from \textsc{HurricaneEmo}. Each sample is annotated with multiple Plutchik-24 emotions.}
\label{tab:dataset_samples}
\end{table}

Table \ref{tab:dataset_stats} presents several statistics of \textsc{HurricaneEmo}. We make three observations. First, the vocabularies across all datasets are large considering there are only 5,000 tweets per hurricane. The vocabularies do decrease by about 30\% after preprocessing, although the resulting sizes still suggest users use a myriad of words to express their emotions. Second, only about 50\% of Harvey tweets and 40\% of Irma/Maria tweets contain hashtags. Hashtags are a unique marker of Twitter discourse \citep{ritter-etal-2011-named}, but in our dataset specifically, hashtags are used to tag particular entities, spread disaster-relief awareness, and create trending content. This phenomena alone makes our tweets different from those collected through distant supervision \citep{abdul-mageed-ungar-2017-emonet}. Third, roughly 80-85\% of tweets contain links to third-party content. Users commonly use links to share news articles, resources for humanitarian aid, and other miscellaneous multimedia.

Table \ref{tab:dataset_samples} shows three samples from \textsc{HurricaneEmo}. Unlike \textsc{EmoNet} \citep{abdul-mageed-ungar-2017-emonet}, our dataset does not have the strong assumption that only one emotion can be expressed in a tweet. For example, the first tweet lexically points towards the expression of more than one emotion. The predicate ``helped us'' implies the user admires Mexico for providing aid, and the exclamation mark is indicative of \joy. In addition, our samples contain a mix of \textit{implicit} and \textit{explicit} emotions, which lexical information alone cannot resolve. In the third tweet, there are no particular words that point towards \anger\sepr and \annoyance, but we can infer the user is upset that the media is not prioritizing Hurricane Maria.

Finally, our emotion prediction tasks cannot be solved by simply retrofitting pre-trained word embeddings \citep{mikolov2013distributed, pennington-etal-2014-glove} or contextualized representations \citep{peters-etal-2018-deep,devlin-etal-2019-bert,liu2019roberta}, which we also empirically show in our experiments (\S\ref{sec:baseline-modeling}). These methods work best for \textit{explicit} emotion detection as they largely overfit to sparse lexical features. Rather, in order to capture \textit{implicit} emotions, models must carry an inductive bias that appropriately reasons over the context (e.g., what event(s) occurred?) and semantic roles (e.g., what happened to whom?) while balancing the aforementioned features.

\begin{table}[t!]
\small
\centering
\begin{tabular}{llll}
\toprule
\multicolumn{2}{c}{Plutchik-8} & \multicolumn{2}{c}{Plutchik-24} \\
\cmidrule(r){1-2} \cmidrule(r){3-4}
\multicolumn{1}{l}{Emotion} & \multicolumn{1}{l}{Abbrv.} & \multicolumn{1}{l}{Emotion} & \multicolumn{1}{l}{Abbrv.} \\
\midrule
\multirow{3}{*}{aggressiveness} & \multirow{3}{*}{\textit{agrsv}} & rage & \textit{rage} \\
 &  & anger & \textit{anger} \\
 &  & annoyance & \textit{anyce} \\ \midrule
\multirow{3}{*}{optimism} & \multirow{3}{*}{\textit{optsm}} & vigilance & \textit{vglnc} \\
 &  & anticipation & \textit{antcp} \\
 &  & interest & \textit{inrst} \\ \midrule
\multirow{3}{*}{love} & \multirow{3}{*}{\textit{love}} & ecstasy & \textit{ecsty} \\
 &  & joy & \textit{joy} \\
 &  & serenity & \textit{srnty} \\ \midrule
\multirow{3}{*}{submission} & \multirow{3}{*}{\textit{sbmsn}} & admiration & \textit{admrn} \\
 &  & trust & \textit{trust} \\
 &  & acceptance & \textit{acptn} \\ \midrule
\multirow{3}{*}{awe} & \multirow{3}{*}{\textit{awe}} & terror & \textit{trror} \\
 &  & fear & \textit{fear} \\
 &  & apprehension & \textit{aprhn} \\ \midrule
\multirow{3}{*}{disapproval} & \multirow{3}{*}{\textit{dspvl}} & amazement & \textit{amzmt} \\
 &  & surprise & \textit{srpse} \\
 &  & distraction & \textit{dstrn} \\ \midrule
\multirow{3}{*}{remorse} & \multirow{3}{*}{\textit{rmrse}} & grief & \textit{grief} \\
 &  & sadness & \textit{sadns} \\
 &  & pensiveness & \textit{psvne} \\ \midrule
\multirow{3}{*}{contempt} & \multirow{3}{*}{\textit{cntmp}} & loathing & \textit{lthng} \\
 &  & disgust & \textit{dsgst} \\
 &  & boredom & \textit{brdom} \\
 \bottomrule
\end{tabular}
\caption{Plutchik-8 (left) and Plutchik-24 (right) abbreviations used throughout this paper.}
\label{tab:emo-codes}
\end{table}

\subsection{Fine-Grained Emotions}
\label{sec:fine-grained-emotions}

We begin to analyze the fine-grained emotions present in our datasets. We ask the following questions: What is the general distribution of emotions? Are certain emotion groups highlighted more than others? How does the distribution change across hurricanes?

\begin{figure}[t!]
    \centering
    \includegraphics[width=7.5cm]{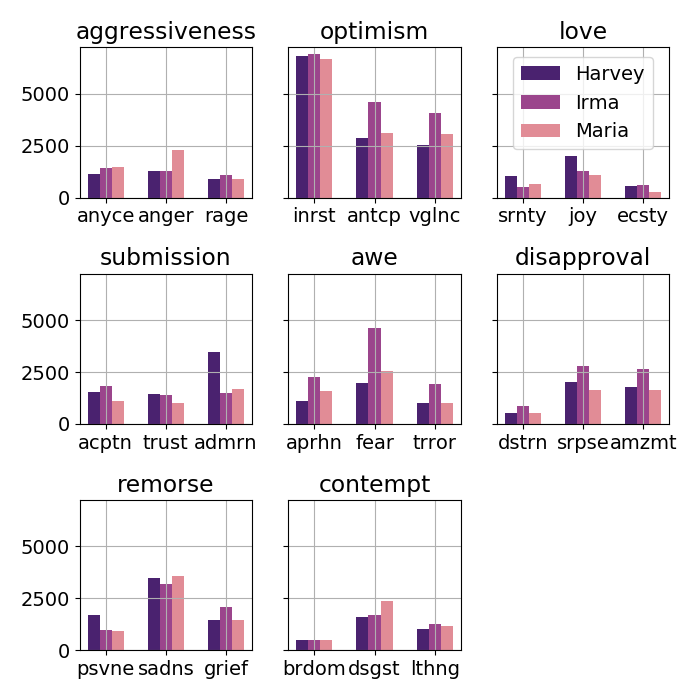}
    \caption{Per-hurricane emotion counts where each box's Plutchik-8 emotion is broken down into its respective Plutchik-24 emotions. Plutchik-24 emotions are abbreviated using the codes in Table \ref{tab:emo-codes}.}
    \label{fig:emotions}
\end{figure}

Figure \ref{fig:emotions} shows Plutchik-24 emotion distributions for Hurricanes Harvey, Irma, and Maria. From these plots, a couple of trends emerge. First, the Plutchik-24 emotion counts are within the ballpark of each other with the notable exceptions of \admiration\sepr and \fear. This suggests that, on average, hurricane disasters evoke a similar spread of implicit and explicit emotions among most emotion categories. Second, users tend to post more optimistic content during hurricane disasters. We hypothesize that users use Twitter as a social platform to spread awareness of the hurricanes themselves or post-disaster relief efforts, commonly using hashtags like \textit{\#prayfortexas}, \textit{\#floridaevacuation}, and \textit{\#donationdrive}. It is encouraging to see that although users do express natural emotions such as fear, sadness, and anger, many seek to help others in the face of adversity. Third, sharp changes in emotion counts between Harvey and Irma may be tied to their history. In the 2017 Atlantic hurricane season, Harvey materialized as a Cat-4 hurricane, and Irma followed around two weeks later as a Cat-5 hurricane.\footnote{Abbreviations for Category-$x$. This refers to the Saffir-Simpson scale for classifying hurricanes based on sustained wind speed, which ranges from 1-5 in order of severity.} Through side-by-side comparisons of both hurricanes' tweets, we found the Irma tweets had more descriptions of destruction and its aftermath. These changes in discourse potentially explain shifts between the emotion distributions.

\begin{figure}[t!]
    \centering
    \includegraphics[width=7.5cm]{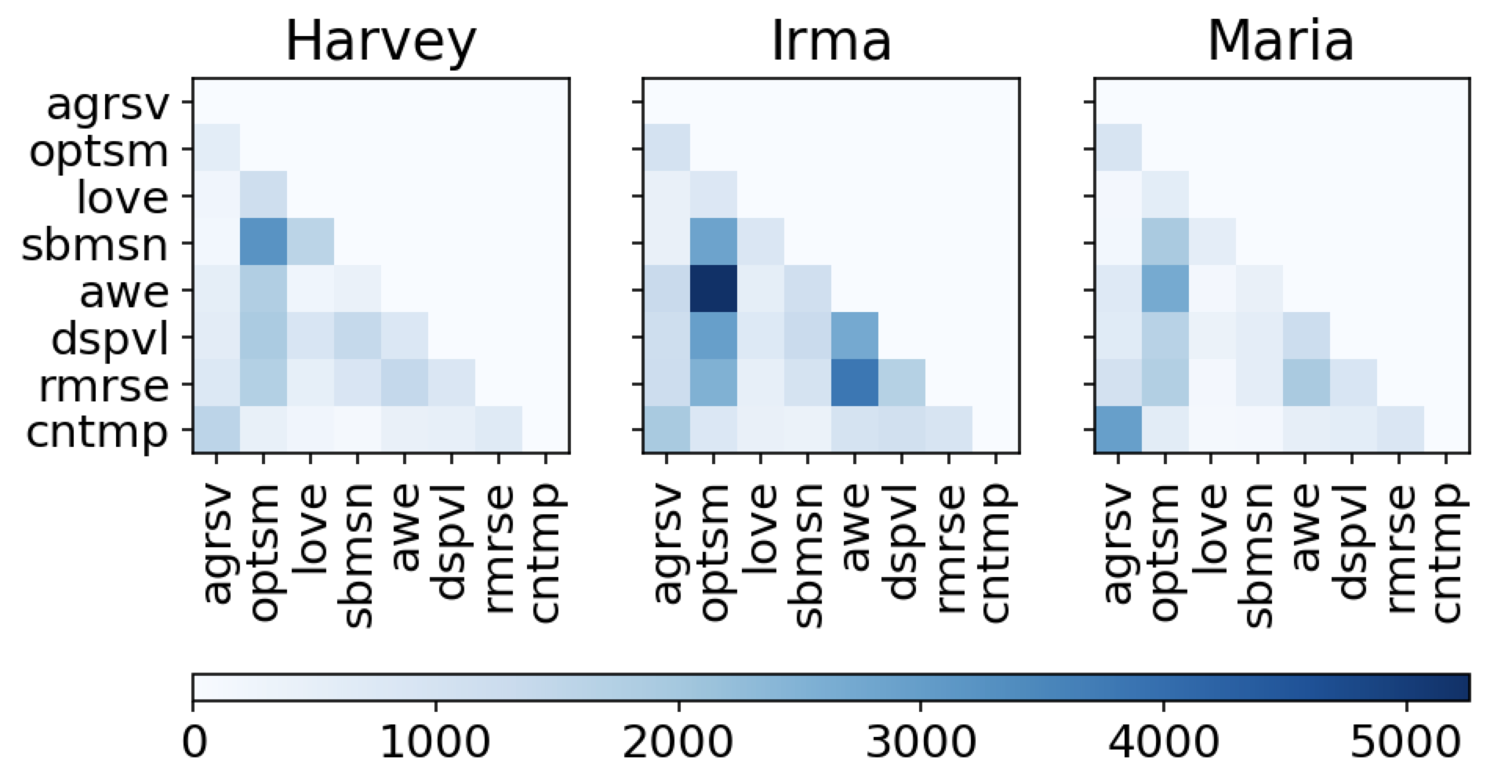}
    \caption{Per-hurricane Plutchik-8 emotion co-occurrences. The matrices are symmetric across the diagonal, so we mask the upper diagonal of the matrix for clarity. Plutchik-8 emotions are abbreviated using the codes in Table \ref{tab:emo-codes}.}
    \label{fig:cooccurrence}
\end{figure}

\subsection{Emotion Co-Occurrence}
\label{sec:emotion-cooccurrence}

Thus far, we have analyzed each Plutchik-24 emotion in isolation. In this section, we ask the following questions: How do Plutchik-8 emotion groups co-occur with one another? Do co-occurrence patterns change across hurricanes?

Figure \ref{fig:cooccurrence} shows co-occurrence heatmaps for each hurricane. Intuitively, we see strong correlations between polarized emotions, that is, emotions categorized as \textit{positive} and \textit{negative}. For example, (\love, \aggressiveness) does not appear as frequently as (\love, \optimism) or (\contempt, \aggressiveness). However, this premise does not always hold; the pairs (\{\disapproval, \remorse\}, \optimism) also co-occur across all hurricanes. Representative of this phenomenon is the tweet: \emph{``I'm raising money for Hurricane Maria Destroyed Everything. Click to Donate: \token{<URL>} via \token{<USER>}.''} The user indicates disapproval towards the hurricane by evoking pathos, but also shows optimism by donating money to a relief effort. Finally, similar to our previous observations (\S\ref{sec:fine-grained-emotions}), we notice an increase in co-occurrence frequencies from Harvey $\rightarrow$ Irma. This increase is, somewhat surprisingly, most apparent with (\awe, \optimism), although (\{\disapproval, \remorse\}, \awe) frequencies also exhibit a noticeable gain. Once again, we posit that users may be expressing their sadness regarding the Cat-4 $\rightarrow$ Cat-5 jump, but at the same time, offering solidarity to those affected by the hurricanes.

\section{Baseline Modeling}
\label{sec:baseline-modeling}

We now turn to modeling the emotions in \textsc{HurricaneEmo}. Because Plutchik-24 emotion counts are heavily imbalanced, we group them into Plutchik-8 emotions and consequently create 8 binary classification tasks. 

\begin{table}[t!]
\small
\centering
\begin{tabular}{lrrr}
\toprule
Plutchik-8 Emotion & Train & Valid & Test \\
\midrule
Aggressiveness & 4,209 & 526 & 527 \\
Optimism & 11,902 & 1,488 & 1,488 \\
Love & 2,569 & 321 & 322 \\
Submission & 6,092 & 762 & 762 \\
Awe & 7,324 & 916 & 916 \\
Disapproval & 5,931 & 741 & 742 \\
Remorse & 7,732 & 967 & 967 \\
Contempt & 3,763 & 470 & 471 \\
\bottomrule
\end{tabular}
\caption{Train, validation, and test splits for each Plutchik-8 emotion.}
\label{tab:splits}
\end{table}

The tweets are assorted into their respective label buckets; because tweets may be labeled with more than one emotion, each belongs to one or more buckets. These buckets represent \textit{positive} samples (i.e., tweets labeled with that emotion). To create \textit{negative} samples, we sample an equal amount from other buckets. From here, we shuffle the positive and negative samples and perform an 80/10/10 split to create the train, validation, and test sets.\footnote{We also experimented with keeping all negative samples as opposed to sampling an equal amount. Each binary task had around 5-7x more negative samples; this significantly hurt model performance. Even with a class imbalance penalty, the models almost never predicted positive samples. Note that although, in aggregate, the number of positive and negative samples match, they do not necessarily match in the train, validation, and test splits.} Table \ref{tab:splits} enumerates the splits.

\begin{table*}[t!]
\small
\centering
\begin{tabular}{rccccccccc}
\toprule
 & AGR & OPT & LOV & SBM & AWE & DSP & RMR & CNT & AVG \\ \midrule
Logistic Reg. & 49.8 & 74.7 & 50.9 & 50.6 & 48.9 & 49.7 & 48.3 & 46.8 & 52.5 \\
Char CNN & 50.2 & 74.3 & 43.0 & 47.2 & 44.7 & 47.1 & 47.4 & 48.8 & 50.3 \\
Word CNN & 43.6 & 74.5 & 44.7 & 45.4 & 44.2 & 47.0 & 46.9 & 43.9 & 48.8 \\
GRU & 48.4 & 74.7 & \textbf{54.0} & 50.9 & 50.1 & 49.9 & 48.9 & 49.2 & 53.3 \\
BERT & \textbf{67.6} & \textbf{75.0} & \textbf{54.0} & \textbf{67.4} & \textbf{68.3} & \textbf{55.7} & \textbf{58.5} & \textbf{66.8} & \textbf{64.1} \\
RoBERTa & 59.7 & 74.7 & \textbf{54.0} & 62.3 & 56.0 & 50.9 & 49.7 & 56.4 & 58.0\\
\bottomrule
\end{tabular}
\caption{Plutchik-8 binary task accuracies, including aggressiveness (agr), optimism (opt), love (lov), submission (sbm), awe (awe), disapproval (dsp), remorse (rmr), contempt (cnt). We also report an average (avg) across all binary tasks. Best results are \textbf{bolded}.}
\label{tab:binary-task-results}
\end{table*}

\subsection{Experimental Setup}
\label{sec:baseline-modeling-setup}

We consider both traditional neural models and pre-trained language models. We implement our models in PyTorch \citep{paszke2019pytorch} and perform all experiments on an NVIDIA Titan V GPU. Training and optimization hyperparameters are detailed in Appendix \ref{sec:appendix-baseline-modeling}. We report mean performance across 10 runs, each with a different random initialization. Below, we elaborate on our models:

\paragraph{Traditional Neural Models.} Each is equipped with 200D GloVe embeddings pre-trained on 2B tweets \citep{pennington-etal-2014-glove}: (1) \textbf{Logistic Regression:} We average the word embeddings of each token in the sequence \citep{iyyer-etal-2015-deep}; (2) \textbf{CNN:} A word-level CNN \citep{kim-2014-convolutional} with 100 filters of size [3, 4, 5] obtains representations. They are max-pooled and concatenated row-wise. We also experiment with a character-level CNN with filter sizes [5, 6, 7]; (3) \textbf{GRU:} A one-layer, unidirectional GRU \citep{cho-etal-2014-learning} with a hidden dimension of 100 obtains features, which are mean pooled. For all models, penultimate representations are projected with a weight matrix $\mathbf{W} \in \mathbb{R}^{d \times 2}$.

\paragraph{Pre-trained Language Models.} We fine-tune base versions of BERT \citep{devlin-etal-2019-bert} and RoBERTa \citep{liu2019roberta} using the HuggingFace Transformers library \citep{wolf2019huggingface}. We use the sentence representations embedded in the \token{[CLS]} token, then project it with a weight matrix $\mathbf{W} \in \mathbb{R}^{d \times 2}$. The language model and classification parameters are jointly fine-tuned.

\subsection{Results}
\label{sec:baseline-modeling-results}

Table \ref{tab:binary-task-results} presents our classification results. We make the following observations:

\paragraph{BERT consistently outperforms other models on most emotion tasks.} BERT shows strong performance across all 8 binary tasks in comparison to traditional neural models and RoBERTa. Unlike most traditional neural models, its accuracy never falls below random chance, showing it captures at least some of the complex phenomena present in our dataset. However, our tasks remain challenging for both types of models alike. For traditional models, word embeddings alone do not provide enough representational power to model our emotional contexts. Although GRUs perform well on \textsc{EmoNet} \citep{abdul-mageed-ungar-2017-emonet}, we suspect that they simply memorize emotion lexicons (\S\ref{sec:dataset-overview}), which is not a notable strategy for capturing implicit emotions. Nevertheless, BERT only obtains an average accuracy of about 64\%. This leaves plenty of room for future work; we perform a comprehensive error analysis as a step towards this goal (\S\ref{sec:baseline-modeling-error-analysis}).

\paragraph{``Better'' pre-trained models (e.g., RoBERTa) do not necessarily help performance.} Unlike popular benchmarks such as GLUE \citep{wang-etal-2018-glue} where more pre-training monotonically increases performance, rather encouragingly, we do not observe the same trend. RoBERTa's average performance is around 5\% better than GRU's, but still around 6\% worse than BERT's. We hypothesize that this drop in performance is attributed to pre-training $\rightarrow$  fine-tuning domain discrepancies. That is, RoBERTa's (additional) pre-training data (e.g., CC-News) may be too distant from Twitter data, which is known for its short contexts and unique vernacular \citep{ritter-etal-2011-named}. We encourage practitioners to avoid applying state-of-the-art models without augmenting them with task-guided pre-training objectives, as we explore later (\S\ref{sec:task-guided-pretraining}).

\begin{table*}[t!]
\small
\centering
\setlength{\tabcolsep}{5pt}
\begin{tabular}{lccccccccc}
\toprule
& AGR & OPT & LOV & SBM & AWE & DSP & RMR & CNT & AVG \\
 \midrule
\textsc{No-Pretrain} & 67.6 & 75.0 & 54.0 & 67.4 & 68.3 & 55.7 & 58.5 & 66.8 & 64.1 \\
\midrule
\multicolumn{10}{l}{\textbf{Supervised Transfer}} \\
\midrule
\textsc{EmoNet} & \inc{39.33}{73.5} & \inc{1.33}{75.2} & \inc{8.00}{55.2} & \inc{9.33}{68.8} & \dec{5.33}{67.5} & \dec{17.33}{53.1} & \inc{10.00}{60.0} & \inc{32.67}{71.7} & \inc{10.00}{65.6} \\
\textsc{Sentiment} & \inc{34.67}{72.8} & \inc{5.33}{75.8} & \inc{58.00}{62.7} & \inc{24.00}{71.0} & \dec{18.00}{65.6} & \dec{15.33}{53.4} & \dec{10.00}{57.0} & \inc{3.33}{67.3} & \inc{10.67}{65.7} \\
\midrule
\multicolumn{10}{l}{\textbf{Unsupervised Transfer}} \\
\midrule
\textsc{EmoNet} & \inc{30.00}{72.1} & \inc{0.67}{75.1} & 54.0 & \dec{42.67}{61.0} & \dec{21.33}{65.1} & \dec{10.00}{54.2} & \inc{14.67}{60.7} & \inc{17.33}{69.4} & \dec{1.33}{63.9} \\
\textsc{Sentiment} & \inc{10.00}{69.1} & \dec{0.67}{74.9} & \dec{2.67}{53.6} & \dec{8.00}{66.2} & \dec{6.67}{67.3} & \dec{9.33}{54.3} & \dec{4.00}{57.9} & \dec{16.00}{64.4} & \dec{4.00}{63.5} \\
\textsc{HurricaneExt} & \inc{40.00}{73.6} & \inc{2.67}{75.4} & \inc{100.00}{69.8} & \inc{10.00}{68.9} & \inc{9.33}{69.7} & \inc{14.67}{57.9} & \inc{11.33}{60.2} & \inc{22.67}{70.2} & \inc{27.33}{68.2} \\
\bottomrule
\end{tabular}
\caption{Task-guided pre-training accuracies (abbreviations defined in Table \ref{tab:binary-task-results}). Displayed in order of supervised (middle) and unsupervised (bottom) pre-training. Results are highlighted with \colorbox{cyan!50}{blue} ($\uparrow$) and \colorbox{red!50}{red} ($\downarrow$) with respect to \textsc{No-Pretrain}. Best viewed in color.}
\label{tab:pretraining-results}
\end{table*}

\subsection{Error Analysis}
\label{sec:baseline-modeling-error-analysis}

Using our BERT model, we sample 25 test errors from each of the 8 emotion tasks, yielding a total of 200 errors. We group the errors into the following categories: lexical and syntactic cues (45\%), insufficient context (24\%), entity mentions (15\%), subjective labeling (10\%), and unknown reasons (6\%). The top three categories are discussed below:

\paragraph{Lexical and Syntactic Cues.} BERT often relies on surface-level lexical features to make predictions, as do most emotion prediction models. This bias also extends to certain syntactic features, such as punctuation. In \emph{``pls be safe everyone!!!!''}, BERT associates the exclamation mark with a positive emotion, but here, the speaker is more concerned.

\paragraph{Insufficient Context.} Users often comment on events, public policies, or linked content that, by themselves, do not carry features for supervised learning. This type of error is not necessarily a shortcoming of BERT, but rather our dataset. For example, in \emph{``for [tracy mcgrady]$_1$, [hall induction]$_2$ muted by effects of [hurricane harvey]$_3$ at home''}, one use external knowledge to reason between the noun phrases and discern the latent emotions.

\paragraph{Entity Mentions.} BERT also makes erroneous predictions in the presence of certain entity mentions. For example, BERT classifies this tweet as \aggressiveness: \emph{``nytimesworld: mexico offered aid to texas after harvey. but after an earthquake and hurricane, it says all help is needed at home.''} Here, the user is merely quoting a news statement as opposed to formulating opinions regarding NY Times' discourse. Because the sentiment towards NY Times is negative in our datasets  overall (due to public backlash on its stories), BERT likely capitalizes on this mention-emotion bias.

\section{Task-Guided Pre-training}
\label{sec:task-guided-pretraining}

To improve upon our baselines, we explore pre-training as a means of \textit{implicitly} incorporating an inductive bias into our BERT model. Our hope is that these pre-training tasks will not only make BERT more robust in the Twitter domain, but also provide useful (albeit abstract) knowledge for the end emotion prediction tasks. For brevity, we chiefly focus on BERT, although our methods can be generalized to other pre-trained models.

\paragraph{Setup.}
\label{sec:task-guided-pretraining-setup}

We explore, in isolation, \textit{supervised} and \textit{unsupervised} pre-training tasks. For the supervised setting, we pre-train on a multi-class emotion task (\textsc{EmoNet}) \citep{abdul-mageed-ungar-2017-emonet} and binary sentiment analysis task (\textsc{Sentiment}) \citep{go2009twitter}. For the unsupervised setting, we pre-train on dynamic masked language modeling \citep{liu2019roberta} on (unlabeled) samples from \textsc{EmoNet}, \textsc{Sentiment}, and \textsc{HurricaneExt} (\S\ref{sec:preprocessing}). For both types of tasks, we further pre-train BERT for a fixed number of epochs, then fine-tune it on a \textsc{HurricaneEmo} task. We compare these results to \textsc{No-Pretrain}, namely the BERT results verbatim from Table \ref{tab:binary-task-results}. We report mean performance across 10 pre-training $\rightarrow$ fine-tuning runs. Further training details, including samples sizes for the pre-training tasks, are available in Appendix \ref{sec:appendix-task-guided-pretraining}.

\paragraph{Results.}

Table \ref{tab:pretraining-results} shows the pre-training results. Supervised pre-training significantly helps with 3-4 emotions, but degrades overall performance on 2-4 emotions. We posit \textsc{Sentiment} aids emotions with highly predictive features. For example, ``wtf'' in \emph{``it's literally the size of texas. wtf''} is correlated with \aggressiveness, but no such lexical cues exist in \emph{``not all heros wear capes <3 thank you stanley - homeless \#hurricane evacuee grooms lost pets,''} which is an \awe\sepr sample.

The unsupervised pre-training results also show a couple trends. First, \textsc{EmoNet} largely hurts downstream performance, especially reducing \submission\sepr accuracy by -6\%. Second, \textsc{Sentiment} (in its \textit{unlabeled} form) yields no noticeable benefits. This implies sentiment information is much more valuable, but of course, subject to the fact that the emotion task is heavily aligned with the original sentiment task. Third, we obtain encouraging results with \textsc{HurricaneExt} pre-training. The gains are most noticeable on \aggressiveness\sepr and \love, but this objective adds +1-2\% accuracy for tasks on which supervised pre-training suffered.

\begin{table*}[t!]
\small
\centering
\setlength{\tabcolsep}{5pt}
\begin{tabular}{rccccccccc}
\toprule
 & AGR & OPT & LOV & SBM & AWE & DSP & RMR & CNT & AVG \\
\midrule
\textsc{Src-Only} & 53.3 & 42.2 & 43.4 & 47.1 & 54.7 & 49.8 & 62.5 & 56.5 & 51.2 \\
\midrule
\textsc{Pretrain-Src} & \inc{10.00}{54.8} & \inc{6.67}{43.2} & \inc{11.33}{45.1} & \inc{4.67}{47.8} & \dec{2.00}{54.4} & \inc{4.00}{50.4} & \inc{5.33}{63.3} & \inc{4.00}{57.1} & \inc{5.33}{52.0} \\
\textsc{Pretrain-Trg} & \inc{11.33}{55.0} & \inc{13.33}{44.2} & \inc{18.67}{46.2} & \inc{6.00}{48.0} & \inc{5.33}{55.5} & \inc{0.67}{49.9} & \inc{8.00}{63.7} & \inc{26.67}{60.5} & \inc{11.33}{52.9} \\
\textsc{Pretrain-Joint} & \dec{4.00}{52.7} & \inc{13.33}{44.2} & \inc{14.00}{45.5} & \inc{4.67}{47.8} & \inc{0.67}{54.8} & \inc{0.67}{49.9} & \dec{6.00}{61.6} & \dec{1.33}{56.3} & \inc{2.67}{51.6} \\
\midrule
\textsc{Trg-Only} & 67.6 & 75.0 & 54.0 & 67.4 & 68.3 & 55.7 & 58.5 & 66.8 & 64.1 \\
\bottomrule
\end{tabular}
\caption{Unsupervised domain adaptation accuracies (abbreviations defined in Table \ref{tab:binary-task-results}). Results are highlighted with \colorbox{cyan!50}{blue} ($\uparrow$) and \colorbox{red!50}{red} ($\downarrow$) with respect to \textsc{Src-Only}. Best viewed in color.}
\label{tab:unsupervised-da-results}
\end{table*}

\section{Fine-Grained Unsupervised Domain Adaptation}
\label{sec:fine-grained-uda}

When new disasters emerge, it is likely we may not have emotion annotations, as alluded to previously (\S\ref{sec:related-work}). Nevertheless, these annotations would be valuable for organizations trying to understand the emotional profile of users during a crisis \cite{fraustino2012social}. In this section, we explore ways to leverage supervision from large-scale emotion datasets (e.g., \textsc{EmoNet} \citep{abdul-mageed-ungar-2017-emonet}) in providing labels for our hurricane emotion datasets. We frame this problem as unsupervised domain adaptation; \textsc{EmoNet} is the \textit{labeled} source domain and our hurricane datasets are the \textit{unlabeled} target domain. Below, we elaborate on our methods.

\paragraph{Framework.} \textsc{EmoNet} was conceived as a multi-class classification task for Plutchik-8 emotions \citep{abdul-mageed-ungar-2017-emonet}. In contrast, we introduce binary classification tasks, one for each Plutchik-8 emotion. We split the \textsc{EmoNet} multi-class task into 8 binary tasks; this creates a one-to-one alignment between each source and target domain task. We separately perform unsupervised domain adaptation for each binary task.

\paragraph{Methods.} We use our BERT model (without task-guided pre-training) as the underlying classifier. Following \citet{han-eisenstein-2019-unsupervised}, we chiefly focus on using strategic pre-training techniques that enable effective transfer between disparate domains. The systems for comparison are: (1) \textsc{Src-Only}: BERT is trained in the source domain and evaluated in the target domain; (2) \textsc{Trg-Only}: BERT is trained and evaluated in the target domain. These results are borrowed verbatim from Table \ref{tab:binary-task-results}; (3) \textsc{Pretrain-*}: BERT undergoes dynamic masked language modeling pre-training using data from domain \textsc{*}, is trained in the source domain, and finally evaluated in the target domain \citep{han-eisenstein-2019-unsupervised}. \textsc{Pretrain-Src} \textit{only} uses pre-training samples from the source domain, \textsc{Pretrain-Trg} \textit{only} uses samples from the target domain, and \textsc{Pretrain-Joint} uses samples from both the source and target domains.\footnote{\textsc{Pretrain-Joint} is conceptually similar to \textsc{AdaptaBert} in \citet{han-eisenstein-2019-unsupervised}, however, we dynamically generate pre-training data \citep{liu2019roberta}.}
We report mean performance across 10 pre-training $\rightarrow$ fine-tuning runs.

\paragraph{Results.} Table \ref{tab:unsupervised-da-results} shows the unsupervised domain adaptation results. Overall, we do not find a significant increase in performance over the \textsc{Src-Only} baseline. Pre-training consistently adds +1\% in average accuracy, but still leaves a large gap between \textsc{Pretrain-Src} and \textsc{Trg-Only}. Regardless, we have a few observations. First, we do not see a (relatively) large increase in performance for \submission, \awe, \disapproval, and \remorse. These emotions may need more explicit strategies to enable domain adaptation. This is also supported by our previous results (\S\ref{sec:task-guided-pretraining}), where we also do not see a (relatively) large benefit from task-guided pre-training. Second, \textsc{Pretrain-Joint} performs worse than both \textsc{Pretrain-Src} and \textsc{Pretrain-Trg}. We posit that, for our emotion tasks, pre-training with a mixture of domains yields a noisier training signal compared to a parameter bias towards the target domain.

\section{Conclusion}
\label{sec:conclusion}

We present \textsc{HurricaneEmo}, an annotated dataset of perceived emotions spanning 15,000 tweets from multiple hurricanes. Tweets are annotated with fine-grained Plutchik-24 emotions, from which we analyze \textit{implicit} and \textit{explicit} emotions and construct Plutchik-8 binary classification tasks. Comprehensive experiments demonstrate our dataset is a challenging benchmark, even for large-scale pre-trained language models. We release our code and datasets as a step towards facilitating research in disaster-centric domains.

\section*{Acknowledgements}

Thanks to Katrin Erk for reviewing an early version of this manuscript, Yasumasa Onoe for discussions on masked language model pre-training, and the anonymous reviewers for their helpful comments. This work was partially supported by the NSF Grants IIS-1850153, IIS-1912887, and IIS-1903963.

\bibliography{acl2020}
\bibliographystyle{acl_natbib}

\clearpage
\appendix

\section{Domain Shifts}
\label{sec:appendix-domain-shifts}

\begin{figure}[t]
    \centering
    \includegraphics[width=7cm]{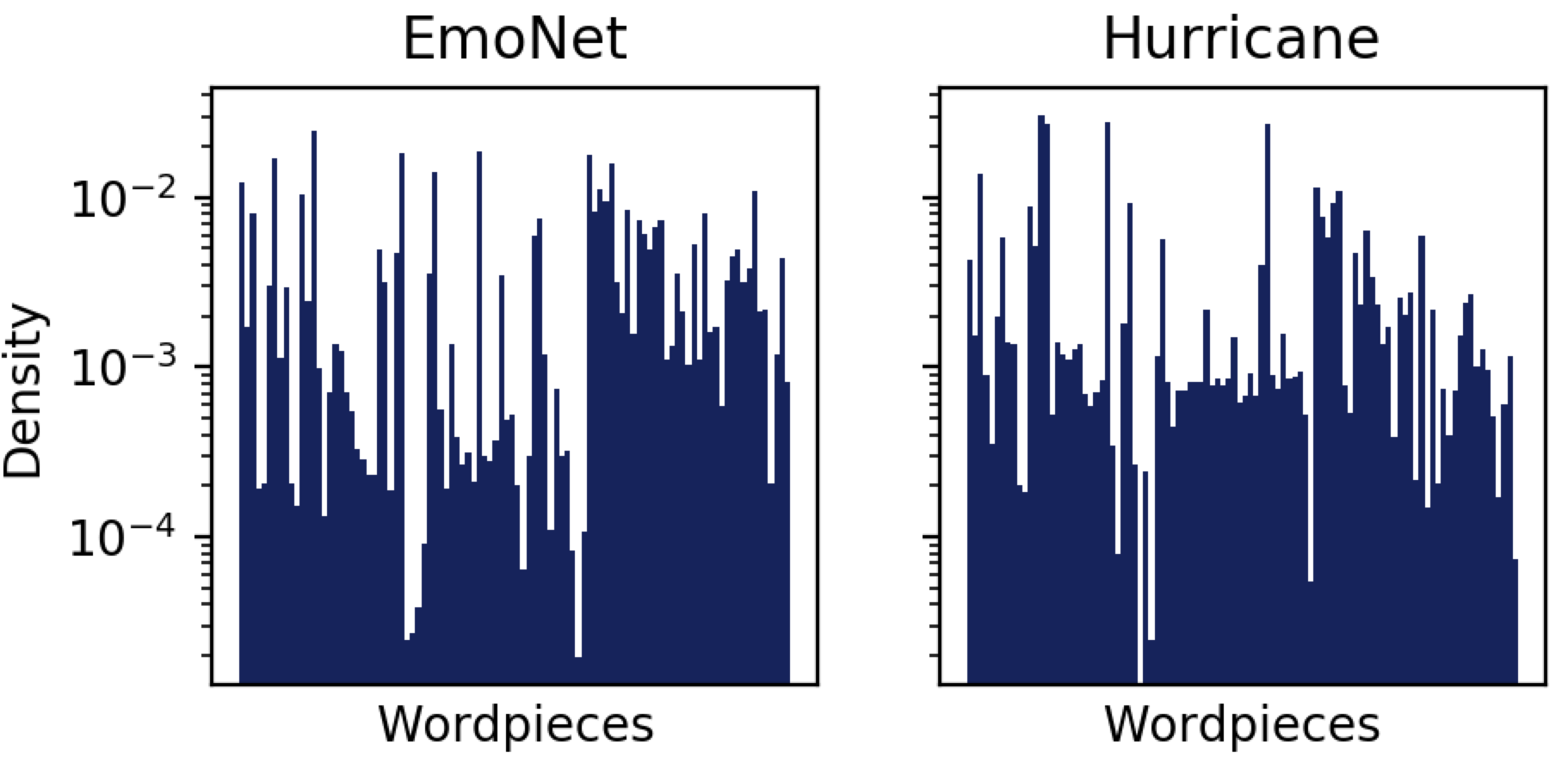}
    \caption{Top 1000 (common) wordpiece densities for \textsc{EmoNet} (left) and \textsc{HurricaneEmo} (right). Densities are calculated by counting wordpiece occurrences and normalizing by the total number of occurrences.}
    \label{fig:appendix-jsd}
\end{figure}

Following the methodology outlined in \citet{desai-etal-2019-evaluating}, we use the Jenson-Shannon Divergence (JSD) between the vocabulary distributions in \textsc{EmoNet} and \textsc{HurricaneEmo} to quantify the domain divergence. The JSD is 0.199, approximately 1e5 larger than those reported in \citet{desai-etal-2019-evaluating}. Figure \ref{fig:appendix-jsd} shows the densities of the top 1000 common wordpieces between both domains. The striking visual differences, even among common wordpieces, indicates a large discrepancy in the input distributions.

\section{Plutchik Emotion Agreement}
\label{sec:appendix-plutchik-emotion-agreement}

\begin{figure*}[t]
    \centering
    \includegraphics[width=8cm]{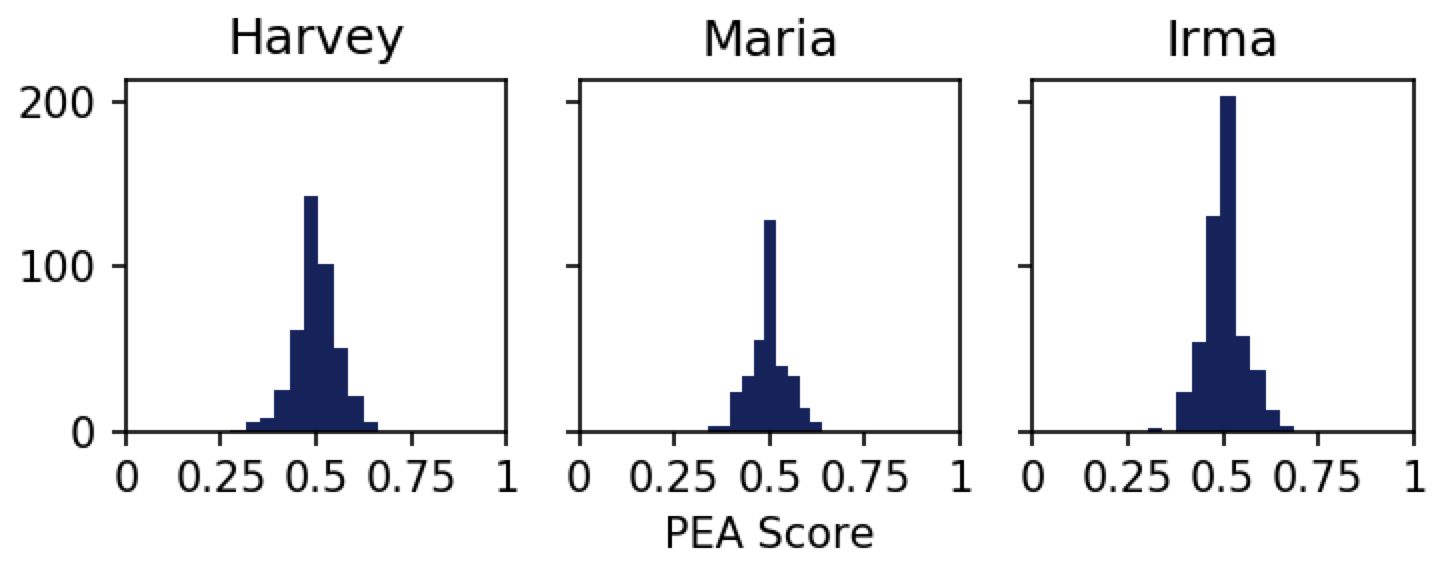}
    \includegraphics[width=8cm]{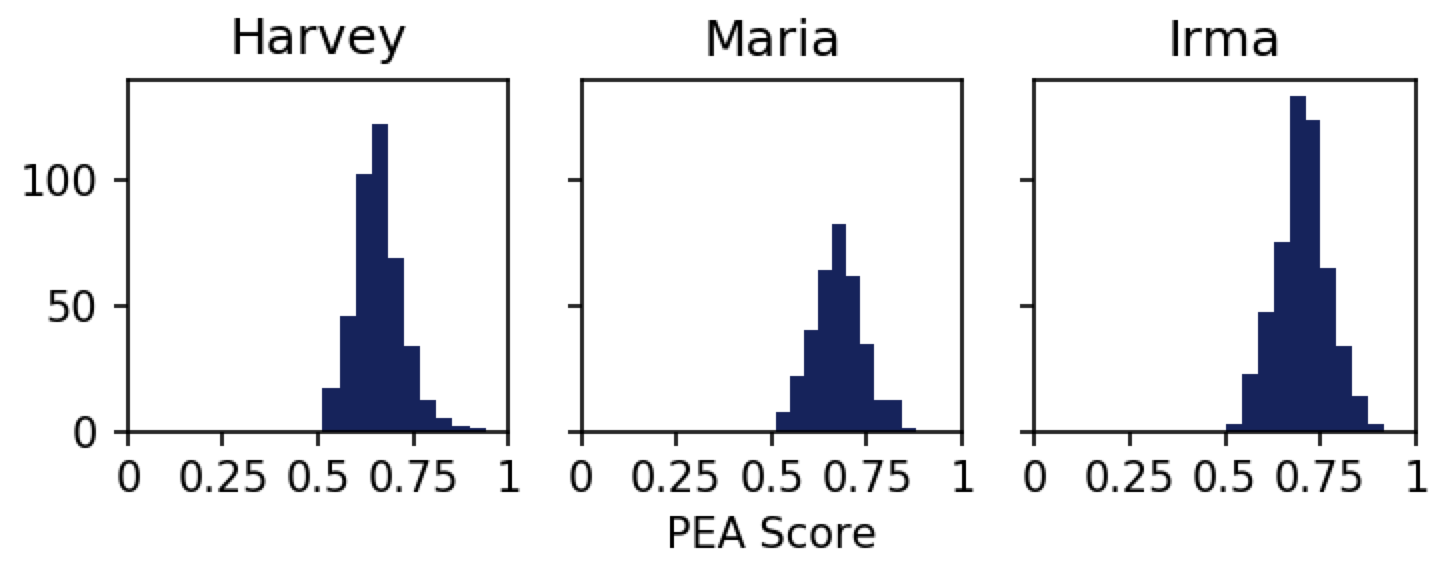}
    \caption{Histograms corresponding to PEA score distributions for random annotations (top) and our annotations (bottom).}
    \label{fig:appendix-pea-experiment}
\end{figure*}

\paragraph{Interpretable Scale.} To assign PEA scores an interpretable scale, we compare randomly generated annotations against our obtained annotations. We detail the process to create random annotations. First, we compute the average number of emotions a worker assigns to a tweet, which evaluates to 3 for all hurricanes. Second, we sample 3 random emotions from the Plutchik-8 wheel for 5000 total annotations. Figure \ref{fig:appendix-pea-experiment} compares the two types of annotations. The per-worker PEA scores for the random annotations collect around the mean (0.5), which is expected due to the law of large numbers. In contrast, the per-worker PEA scores for our annotations are shifted towards the right, indicating better agreement than the random baseline. Therefore, we interpret our annotations as showing ``moderate agreement'' under the PEA metric.

\paragraph{Human Evaluation.} Using our worker annotations across all three hurricanes, we create two annotation pairs for three workers, that is, A: $(w_1, w_2)$ and B: $(w_1, w_3)$, where A and B have a shared worker $w_1$. This format lends a total of 73,418 A/B total pairs. We sample 500 A/B pairs from this pool, initialize each HIT with 10 pairs, and assign 5 total workers per HIT.

\section{Baseline Modeling}
\label{sec:appendix-baseline-modeling}

Table \ref{tab:appendix-base-hyperparams} shows the hyperparameters. For our pre-trained models (e.g., BERT and RoBERTa), we use the default dropout rate (0.1) on the self-attention layers, but do not use additional dropout on the top linear layer. Furthermore, we use gradient accumulation to enable training with larger mini-batches.

\begin{table*}[t]
\small
\centering
\begin{tabular}{lcccccc}
\toprule
 & Logistic Reg. & Word CNN & Char CNN & GRU & BERT & RoBERTa \\
\midrule
Epochs & 5 & 5 & 5 & 5 & 3 & 3 \\
Batch Size & 64 & 64 & 64 & 64 & 16 & 16 \\
Learning Rate & 1e-4 & 1e-3 & 5e-5 & 1e-4 & 2e-5 & 2e-5 \\
Weight Decay & 0 & 0 & 0 & 0 & 0 & 1e-3 \\
Dropout & 0 & 0.5 & 0.7 & 0.7 & -- & -- \\
\bottomrule
\end{tabular}
\caption{Hyperparameters for the baseline modeling experiments (\S\ref{sec:baseline-modeling}).}
\label{tab:appendix-base-hyperparams}
\end{table*}

\section{Task-Guided Pre-training}
\label{sec:appendix-task-guided-pretraining}

\paragraph{Masked Language Modeling.} Following \citet{devlin-etal-2019-bert}, we select 15\% of inputs uniformly at random (except for \token{[CLS]} and \token{[SEP]}) as prediction targets for the masked language modeling task. From the corresponding inputs, 80\% are set to \token{[MASK]}, 10\% are set to random tokens, and 10\% are set to the original tokens. However, we follow \citet{liu2019roberta} in creating pre-training data dynamically, rather than statically. This merely leads to slower convergence times as it becomes more difficult to fit the data. We fine-tune on the pre-training data for 10 epochs using a batch size of 16 and learning rate of 2e-5. Once pre-training concludes, we initialize a BERT model with these weights and fine-tune it on our emotion tasks using the hyperparameters in Table \ref{tab:appendix-base-hyperparams} with a learning rate of 3e-5.

\paragraph{Pre-training Corpus.} Our pre-training corpus is created by concatenating a collection of (shuffled) tweets $x_1, x_2, \cdots, x_n$ together, each separated by \token{[SEP]}. The corpus is split into segments of size 512 with \token{[CLS]} prepended to each one. For clarity, each batch consisting of tokens $x_i, \cdots, x_j$ is constructed as \token{[CLS]} $x_i$ \token{[SEP]} $\cdots$ \token{[SEP]} $x_j$ \token{[SEP]}. We elaborate on two design decisions. First, prepending \token{[CLS]} to each batch, as opposed to each tweet, leads to better results. Second, largely due to computational reasons, we pack disparate tweets together in the same batch.

\section{Extended Pre-training Experiments}
\label{sec:appendix-extended-pretraining-experiments}

\subsection{EmoNet Binary Task Pre-training}

In Section \ref{sec:task-guided-pretraining}, we pre-trained on a \textsc{EmoNet} multi-class classification task. In this section, we explore a fine-grained pre-training scheme. We create Plutchik-8 binary tasks from \textsc{EmoNet}, then fine-tune each emotion model separately on their respective \textsc{HurricaneEmo} tasks. Table \ref{tab:appendix-emonet-binary} shows the results. \textsc{EmoNet-Binary} performs markedly worse than \textsc{EmoNet-Multi} and leads to a -2\% reduction in averaged accuracy. Therefore, multi-class pre-training creates better representations for downstream evaluation, although they are still not as effective as other pre-training methods (e.g., masked language modeling).

\subsection{Varying Amounts of Pre-training Data}

The \textsc{Sentiment} and \textsc{HurricaneExt} datasets contain significantly more samples than currently used. In this section, we study the effects of using varying amounts of pre-training data on downstream \textsc{HurricaneEmo} performance. For both pre-training datasets, we use 1.6M samples. Table \ref{tab:appendix-supervised-pretraining-results} shows the supervised \textsc{Sentiment} results. Tables \ref{tab:appendix-unsup-sentiment-pretraining-results} and \ref{tab:appendix-unsup-disaster-pretraining-results} show the unsupervised \textsc{Sentiment} and \textsc{HurricaneExt} results, respectively. For both types of pre-training tasks, there is no noticeable benefit to using more pre-training data. The supervised \textsc{Sentiment} and unsupervised \textsc{HurricaneExt} results both saturate around 200K samples, which is what we report in our paper. The results for unsupervised \textsc{HurricaneExt} pre-training are especially compelling because they show that, without any labeled data, we can achieve strong downstream results. Finally, the unsupervised \textsc{Sentiment} task yields almost no gains for most emotions, showing that the type of data used for masked language modeling matters. Through side-by-side comparisons, we notice that the \textsc{Sentiment} samples are shorter in length and the \textsc{HurricaneExt} samples contain more relevant content, such as hurricane-specific hashtags.

\begin{table*}[t]
\small
\centering
\setlength{\tabcolsep}{5pt}
\begin{tabular}{rccccccccc}
\toprule
& AGR & OPT & LOV & SBM & AWE & DSP & RMR & CNT & AVG \\
 \midrule
\textsc{No-Pretrain} & 67.6 & 75.0 & 54.0 & 67.4 & 68.3 & 55.7 & 58.5 & 66.8 & 64.1 \\
\midrule
Multi & \inc{39.33}{73.5} & \inc{1.33}{75.2} & \inc{8.00}{55.2} & \inc{9.33}{68.8} & \dec{5.33}{67.5} & \dec{17.33}{53.1} & \inc{10.00}{60.0} & \inc{32.67}{71.7} & \inc{10.00}{65.6} \\
Binary & \inc{0.67}{67.7} & \dec{0.67}{74.9} & \dec{2.00}{53.7} & \dec{18.00}{64.7} & \dec{5.33}{67.5} & \dec{8.00}{54.5} & \dec{18.00}{55.8} & \dec{21.33}{63.6} & \dec{8.67}{62.8} \\
\bottomrule
\end{tabular}
\caption{Pre-training using multi-class and binary \textsc{EmoNet} tasks. See Table \ref{tab:pretraining-results} for styling considerations.}
\label{tab:appendix-emonet-binary}
\end{table*}

\begin{table*}[t]
\small
\centering
\setlength{\tabcolsep}{5pt}
\begin{tabular}{rccccccccc}
\toprule
 & AGR & OPT & LOV & SBM & AWE & DSP & RMR & CNT & AVG \\
\midrule
\textsc{No-Pretrain} & 67.6 & 75.0 & 54.0 & 67.4 & 68.3 & 55.7 & 58.5 & 66.8 & 64.1 \\
\midrule
50 K & \inc{39.33}{73.5} & \inc{2.00}{75.3} & \inc{44.67}{60.7} & \inc{15.33}{69.7} & \dec{8.00}{67.1} & \dec{29.33}{51.3} & \dec{22.00}{55.2} & \dec{3.33}{66.3} & \inc{5.33}{64.9} \\
100 K & \inc{34.67}{72.8} & \inc{5.33}{75.8} & \inc{58.00}{62.7} & \inc{24.00}{71.0} & \dec{18.00}{65.6} & \dec{15.33}{53.4} & \dec{10.00}{57.0} & \inc{3.33}{67.3} & \inc{10.67}{65.7} \\
200 K & \inc{38.67}{73.4} & \inc{4.00}{75.6} & \inc{100.00}{69.1} & \inc{16.00}{69.8} & \dec{12.00}{66.5} & \dec{16.00}{53.3} & \dec{9.33}{57.1} & \inc{20.00}{69.8} & \inc{18.00}{66.8} \\
400 K & \inc{36.67}{73.1} & \inc{2.67}{75.4} & \inc{88.00}{67.2} & \inc{18.00}{70.1} & \dec{17.33}{65.7} & \dec{16.67}{53.2} & \dec{8.67}{57.2} & \inc{4.00}{67.4} & \inc{14.00}{66.2} \\
800 K & \inc{39.33}{73.5} & \inc{2.00}{75.3} & \inc{14.67}{56.2} & \inc{13.33}{69.4} & \dec{21.33}{65.1} & \dec{8.67}{54.4} & \dec{9.33}{57.1} & \inc{9.33}{68.2} & \inc{5.33}{64.9} \\
1600 K & \inc{24.00}{71.2} & \inc{1.33}{75.2} & \inc{72.00}{64.8} & \inc{9.33}{68.8} & \dec{24.00}{64.7} & \dec{4.00}{55.1} & \dec{16.00}{56.1} & \inc{26.00}{70.7} & \inc{11.33}{65.8} \\
\bottomrule
\end{tabular}
\caption{Pre-training using 50-1600K labeled samples from \textsc{Sentiment}. See Table \ref{tab:pretraining-results} for styling considerations.}
\label{tab:appendix-supervised-pretraining-results}
\end{table*}

\begin{table*}[t]
\small
\centering
\setlength{\tabcolsep}{5pt}
\begin{tabular}{rccccccccc}
\toprule
 & AGR & OPT & LOV & SBM & AWE & DSP & RMR & CNT & AVG \\
\midrule
\textsc{No-Pretrain} & 67.6 & 75.0 & 54.0 & 67.4 & 68.3 & 55.7 & 58.5 & 66.8 & 64.1 \\
\midrule
50 K & \inc{20.67}{70.7} & \dec{0.67}{74.9} & \inc{4.00}{54.6} & \dec{7.33}{66.3} & \dec{8.67}{67.0} & \dec{12.00}{53.9} & \inc{5.33}{59.3} & \dec{6.67}{65.8} & \dec{0.67}{64.0} \\
100 K & \inc{26.67}{71.6} & 75.0 & 54.0 & \dec{7.33}{66.3} & \inc{2.00}{68.6} & \dec{4.00}{55.1} & \dec{7.33}{57.4} & \dec{30.00}{62.3} & \dec{2.00}{63.8} \\
200 K & \inc{10.00}{69.1} & \dec{0.67}{74.9} & \dec{2.67}{53.6} & \dec{8.00}{66.2} & \dec{6.67}{67.3} & \dec{9.33}{54.3} & \dec{4.00}{57.9} & \dec{16.00}{64.4} & \dec{4.00}{63.5} \\
400 K & \inc{16.00}{70.0} & \dec{0.67}{74.9} & \dec{1.33}{53.8} & \inc{10.67}{69.0} & \inc{3.33}{68.8} & \dec{8.00}{54.5} & \inc{10.67}{60.1} & \dec{15.33}{64.5} & \inc{2.67}{64.5} \\
800 K & \inc{19.33}{70.5} & \dec{0.67}{74.9} & \inc{7.33}{55.1} & \dec{8.00}{66.2} & \inc{4.67}{69.0} & \dec{16.00}{53.3} & \inc{6.00}{59.4} & \dec{22.67}{63.4} & \dec{0.67}{64.0} \\
1600 K & \inc{10.00}{69.1} & \dec{0.67}{74.9} & \inc{8.67}{55.3} & \dec{6.00}{66.5} & \dec{7.33}{67.2} & \dec{7.33}{54.6} & \inc{5.33}{59.3} & \dec{12.00}{65.0} & \dec{0.67}{64.0} \\
\bottomrule
\end{tabular}
\caption{Pre-training using 50-1600K unlabeled samples from \textsc{Sentiment}. See Table \ref{tab:pretraining-results} for styling considerations.}
\label{tab:appendix-unsup-sentiment-pretraining-results}
\end{table*}

\begin{table*}[t]
\small
\centering
\setlength{\tabcolsep}{5pt}
\begin{tabular}{rccccccccc}
\toprule
 & AGR & OPT & LOV & SBM & AWE & DSP & RMR & CNT & AVG \\
\midrule
\textsc{No-Pretrain} & 67.6 & 75.0 & 54.0 & 67.4 & 68.3 & 55.7 & 58.5 & 66.8 & 64.1 \\
\midrule
50 K & \inc{34.00}{72.7} & 75.0 & \inc{40.00}{60.0} & \dec{1.33}{67.2} & \inc{4.67}{69.0} & \inc{4.67}{56.4} & \inc{12.67}{60.4} & \inc{36.00}{72.2} & \inc{16.67}{66.6} \\
100 K & \inc{28.00}{71.8} & \inc{0.67}{75.1} & \inc{22.67}{57.4} & \inc{11.33}{69.1} & \inc{13.33}{70.3} & \dec{3.33}{55.2} & \inc{26.00}{62.4} & \dec{10.00}{65.3} & \inc{11.33}{65.8} \\
200 K & \inc{40.00}{73.6} & \inc{2.67}{75.4} & \inc{100.00}{69.8} & \inc{10.00}{68.9} & \inc{9.33}{69.7} & \inc{14.67}{57.9} & \inc{11.33}{60.2} & \inc{22.67}{70.2} & \inc{27.33}{68.2} \\
400 K & \inc{25.33}{71.4} & \inc{1.33}{75.2} & \inc{38.00}{59.7} & \inc{15.33}{69.7} & \inc{3.33}{68.8} & \dec{3.33}{55.2} & \inc{14.67}{60.7} & \dec{21.33}{63.6} & \inc{9.33}{65.5} \\
800 K & \inc{25.33}{71.4} & \inc{2.00}{75.3} & \inc{32.67}{58.9} & \inc{13.33}{69.4} & \inc{8.67}{69.6} & \dec{11.33}{54.0} & \inc{12.00}{60.3} & \inc{30.00}{71.3} & \inc{14.67}{66.3} \\
1600 K & \inc{38.00}{73.3} & \inc{4.67}{75.7} & \dec{22.00}{50.7} & \inc{6.00}{68.3} & \dec{18.67}{65.5} & \inc{0.67}{55.8} & \inc{16.67}{61.0} & \dec{18.00}{64.1} & \inc{1.33}{64.3} \\
\bottomrule
\end{tabular}
\caption{Pre-training using 50-1600K unlabeled samples from \textsc{HurricaneExt}. See Table \ref{tab:pretraining-results} for styling considerations.}
\label{tab:appendix-unsup-disaster-pretraining-results}
\end{table*}

\label{sec:appendix-varying-data}

\begin{figure*}
    \centering
    \includegraphics[width=10cm]{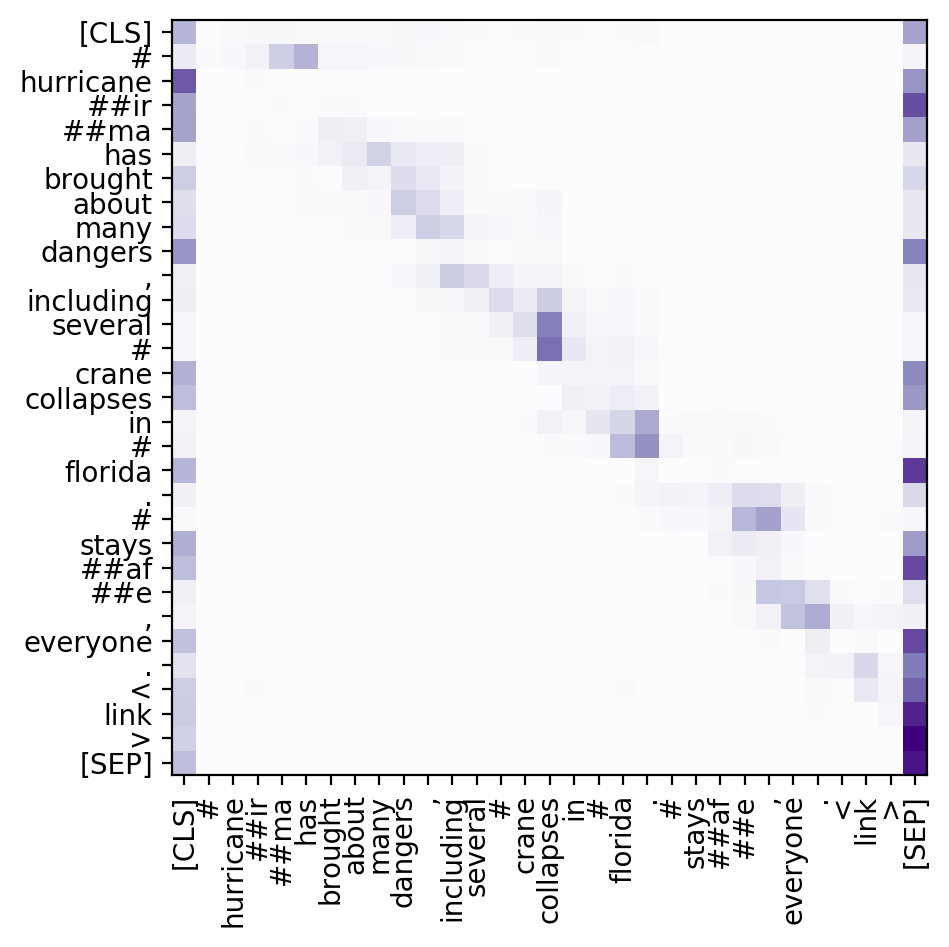}
    \caption{Visualization of BERT's self-attention on a Hurricane Irma sample. In particular, this head captures the entities ``hurricane irma,'' ``florida,'' ``everyone'' and the verb phrase ``crane collapses.''}
    \label{fig:selfatt}
\end{figure*}

\end{document}